\documentclass{article}
\usepackage{spconf,amsmath,graphicx,booktabs,url}
\usepackage{balance}
\usepackage[hidelinks]{hyperref}

\usepackage{iftex}
\ifpdftex
  \usepackage[utf8]{inputenc}
  \usepackage[whole,ipaex-type1]{bxcjkjatype}  %
  \newcommand{\ja}[1]{{\CJKfamily{ipxm}#1}}
\else
  \usepackage{fontspec}
  \newfontfamily\jpfont[Path=./, Scale=0.92]{ipaexm.ttf}
  \newcommand{\ja}[1]{{\jpfont #1}}
\fi
\usepackage[table]{xcolor}

\title{Training-Free Pronunciation Transcription via Text-Constrained Acoustic Rescoring}

\name{Hikaru Asano$^{1,2}$, Yotaro Kubo$^{2}$, So Kuroki$^{2}$}
\address{$^{1}$The University of Tokyo\quad $^{2}$Sakana AI\thanks{This work was done during Hikaru Asano's internship at Sakana AI.}}

\begin{document}
\ninept
\maketitle

\begin{abstract}
  Accurate and efficient pronunciation transcription is essential for preparing text-to-speech training data at scale. Existing approaches have different limitations: grapheme-to-pronunciation (G2P) and speech-to-pronunciation (S2P) methods each capture only partial information, using only text or only speech, while speech-and-text-to-pronunciation (ST2P) methods use both but require costly pronunciation-annotated data. To address this problem, we propose a training-free ST2P pipeline that integrates both lexical and acoustic information at inference time. Lexical resources and G2P tools generate text-constrained candidates, and a left-to-right greedy search selects the best one using whole-sequence negative log-likelihoods from frozen pretrained S2P models. On three Japanese corpora, our method reduces Character Error Rate (CER) from 0.60--1.40\% (text-only baseline) to 0.04--0.17\% with reference transcripts, and 0.64--1.58\% with ASR transcripts. It outperforms all baselines, including a trained ST2P model and commercial multimodal LLMs. Our greedy search method is 3--3.5$\times$ faster than beam search at similar CER, and the cascade is 2$\times$ faster than direct decoding ensuring the efficiency and accuracy. In Spanish, French, and preliminary English, it also surpasses four open multimodal LLMs and the best traditional methods.
\end{abstract}


\section{Introduction}
\label{sec:intro}

Pronunciation transcription is a technique that recovers pronunciation symbols from speech signals or transcribed texts~\cite{Waibel1989-cj,Graves2013-sh}.
It is widely used in diverse applications including training data preparation for text-to-speech (TTS) models, language education, and study of low-resourced languages~\cite{kominek04b_ssw,franco99_eurospeech}.
In particular, keeping pronunciation transcription inexpensive while maintaining low error rates is crucial for preparing TTS training data at scale~\cite{Kim2021-eh,sarashina2026tts}.

\begin{figure}[t]
    \centering
    \includegraphics[width=\columnwidth]{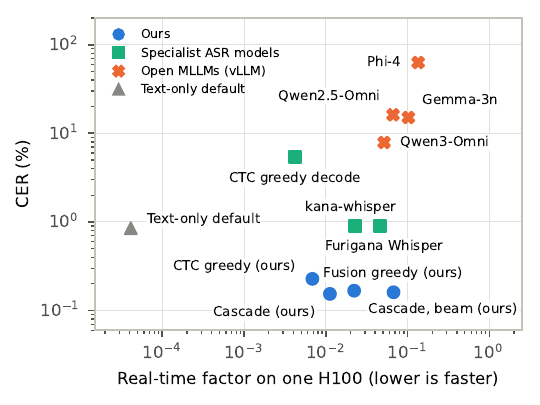}
    \caption{Processing cost versus pronunciation error rates on JVS-dev. The horizontal axis is the real-time factor (log scale; lower is faster) and the vertical axis is character error rate (CER) (log scale; lower is better). The proposed methods (ours) achieves the lowest CER at low computational cost.}
    \label{fig:rtf}
\end{figure}

Two conventional approaches address this task using different inputs.
Grapheme-to-pronunciation (G2P) conversion~\cite{zhu2022charsiu,koriyama2026g2p} predicts a pronunciation sequence $y$ from an orthographic transcript $w$.
Speech-to-pronunciation (S2P) conversion~\cite{bharadwaj2026phoneticxeus,li2026powsm} instead predicts pronunciation from a speech signal $x$.
For the application of TTS data preparation, it is common that both graphemes and speech signals are available for pronunciation prediction.
However, both G2P and S2P approaches are insufficient because they discard information from the other clue for pronunciation transcription.

Speech-and-text-to-pronunciation (ST2P) models address this limitation by jointly using speech and text to predict pronunciation, corresponding to the mapping $(x, w) \to y$.
Related approaches incorporate text through additional text embeddings in an S2P model~\cite{hien2025grapheme} or introduce textual supervision through multitask training that combines speech-to-text and S2P objectives~\cite{kubo2026building}.
Despite their effectiveness, they require pronunciation-annotated training data, which is typically more costly than ordinary text transcription and limits scalability~\cite{Garofolo1993-om}.

\begin{figure*}[t]
    \centering
    \includegraphics[width=\textwidth]{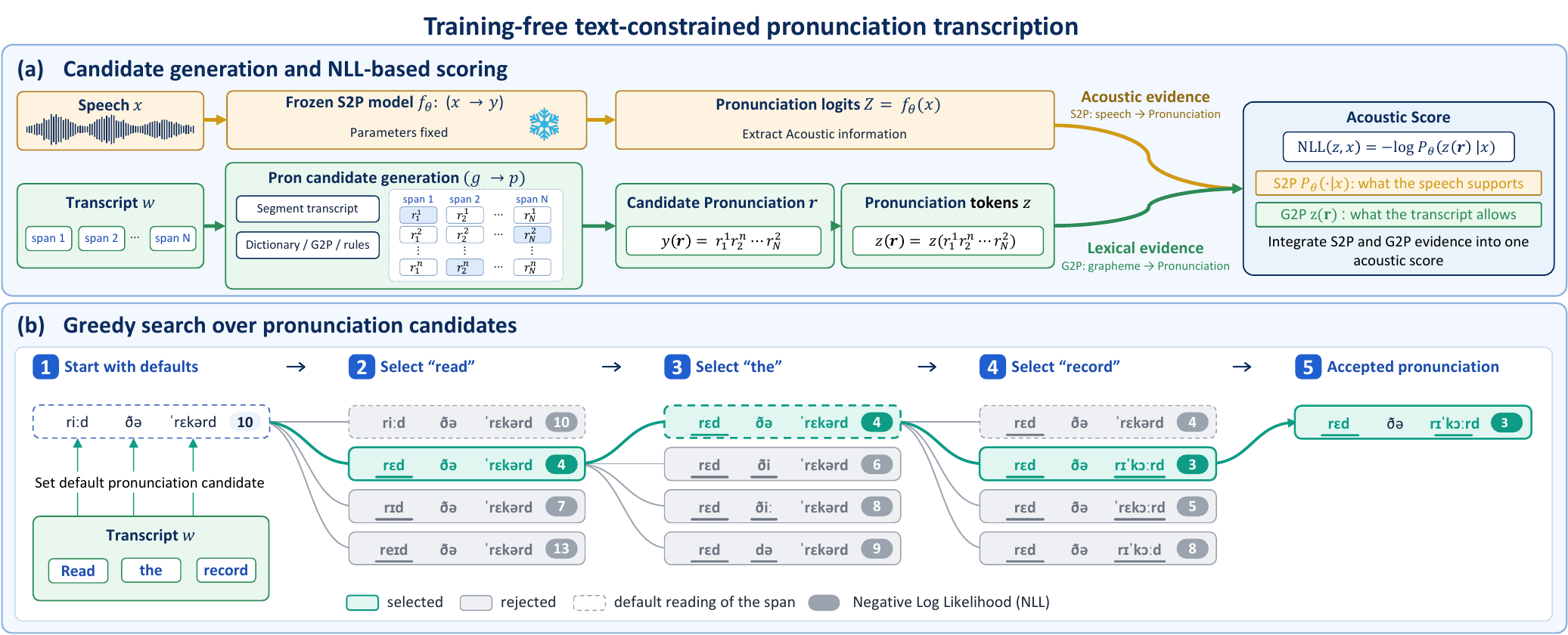}
    \caption{Training-free pronunciation transcription. (a) Dictionaries, G2P, and rules generate readings for transcript spans. Complete pronunciations $y(\mathbf r)$ are converted to model-specific tokens and evaluated by a frozen S2P model given speech $x$, combining lexical constraints and acoustic evidence through negative log-likelihood (NLL). (b) Left-to-right greedy search for ``read the record'' ($P=0$; illustrative NLLs). Each node represents a complete pronunciation with earlier choices fixed and later spans at their defaults. The lowest-NLL candidate (green) is committed; gray nodes are rejected, and dashed outlines mark span defaults. The final node is the search output; cascade rescoring and the subsequent margin check are not shown.}
    \label{fig:pipe}
\end{figure*}

In this paper, we instead explore a ``training-free'' ST2P pipeline that leverages existing lexical resources and G2P tools, together with open pretrained S2P models.
The key idea is to generate text-constrained pronunciation candidates for $w$ from lexical resources and G2P tools, and to select the best one by greedy search using the acoustic scores of frozen S2P models on $x$ (Fig.~\ref{fig:pipe}).
Both clues are thus integrated purely at inference time: the pipeline requires neither audio--text--pronunciation $(x, w, y)$ triplets nor any model training, and is therefore inexpensive while achieving low transcription error rates.

Our contributions are:
(i) We propose a training-free ST2P pipeline that combines lexical candidates with frozen S2P models (Sec.~\ref{sec:method}).
(ii) On three Japanese corpora, our method achieves lower character error rates (CER) than all evaluated baselines, using both reference and ASR transcripts, while maintaining fast inference as shown in Fig.~\ref{fig:rtf} (Sec.~\ref{sec:results}, \ref{sec:cost}).
(iii) We also demonstrate improvements on Spanish, French, and preliminary English, outperforming four open MLLMs and text-only specialists with the best configuration for each language (Sec.~\ref{sec:multilingual}).

\section{Task and training-free setting}
\label{sec:task}

\noindent\textbf{Task.} Given speech $x$ and its transcript $w$, we estimate the pronunciation sequence $y$ actually realized in $x$, rather than its canonical reading. The transcript may come from reference text or ASR output; the latter enables an automatic pipeline.
The pronunciation alphabet matches the language and available resources (e.g., kana for Japanese, IPA otherwise).

\section{Method}
\label{sec:method}

Fig.~\ref{fig:pipe} shows the training-free ST2P pipeline. Candidate pronunciations $y$ for transcript $w$ are generated from lexical resources and scored by frozen S2P models to yield a single acoustic score integrating acoustic and lexical cues (Fig.~\ref{fig:pipe}(a)). The best combination of pronunciation candidates is found by a left-to-right greedy search minimizing negative log-likelihood (NLL) under these models given the speech $x$ (Fig.~\ref{fig:pipe}(b)).

\subsection{Candidates and frozen acoustic scores}
\label{ssec:cand}
As in Fig.~\ref{fig:pipe}(a), the transcript $w$ is segmented into $N$ spans, e.g., by morphological analysis. For each span $n$, lexical resources such as dictionaries, G2P, or rules provide a finite set $\mathcal{R}_n$ of candidate readings. One of them, the default reading $r_n^0 \in \mathcal{R}_n$, is the most probable reading predicted from the text alone. Choosing one reading $r_n \in \mathcal{R}_n$ for every span gives an assignment $\mathbf{r} = (r_1, \ldots, r_N) \in \mathcal{R} = \prod_n \mathcal{R}_n$, whose complete pronunciation is the concatenation
\begin{equation}
 y(\mathbf r)=r_1\,r_2\cdots r_N .
 \label{eq:complete}
\end{equation}
Instead of scoring each span separately, we evaluate $y(\mathbf r)$ as a whole on the audio $x$, jointly judging all readings in context, and seek
\begin{equation}
 \hat{\mathbf r}=\operatorname*{arg\,min}_{\mathbf r\in\mathcal R} J(\mathbf r,x),\quad
 J(\mathbf r,x)=S(y(\mathbf r),x)+P(\mathbf r),
 \label{eq:objective}
\end{equation}
where $S$ is the acoustic score, and the optional penalty $P$ favors more reliable candidates in ambiguous cases.
The acoustic score is the NLL of the candidate under a frozen S2P model,
\begin{equation}
 S(y,x)=-\log p(z(y)\mid x),
 \label{eq:nll}
\end{equation}
where $z$ renders $y$ into the alphabet and tokens of the model.

\subsection{Greedy search and cascade}
\label{ssec:gate}
As $|\mathcal R|$ grows exponentially with $N$, we approximate the objective in Eq.~\eqref{eq:objective} by a left-to-right greedy search (Fig.~\ref{fig:pipe}(b)), assuming each span's acoustic evidence is mostly local.
Let $\mathbf r[n\!\leftarrow\! r]$ denote $\mathbf r$ with its $n$-th span replaced by $r$. Starting from $\hat{\mathbf r}^{(0)}=\mathbf r^0=(r_1^0,\ldots,r_N^0)$, we visit each span once. At span $n$, we select the best candidate and commit it:
\begin{equation}
 r_n'=\operatorname*{arg\min}_{r\in\mathcal R_n}
 J(\hat{\mathbf r}^{(n-1)}[n\!\leftarrow\! r],x),\;\;
 \hat{\mathbf r}^{(n)}=\hat{\mathbf r}^{(n-1)}[n\!\leftarrow\! r_n'],
 \label{eq:greedy}
\end{equation}
so that $\hat{\mathbf r}^{(n)}=(r_1',\ldots,r_n',r_{n+1}^0,\ldots,r_N^0)$. Thus, the $|\mathcal R_n|$ candidates are scored as complete sequences, with earlier choices fixed and later spans at their defaults.
The final assignment is $\hat{\mathbf r}=\hat{\mathbf r}^{(N)}$.

\noindent\textbf{Second scorer and cascade.}
For added robustness, we can combine two frozen S2P models using the fused score $S=L_1+\lambda L_2$, where $L_i$ is the NLL from model $i$ and $\lambda\ge 0$ controls the second model's weight. To reduce computation, we use a cascade: we first run the greedy search with $S=L_1$. If $\lambda>0$ and this pass changes any default reading, we rerun the search from $\mathbf r^0$ with the fused score; otherwise, we retain the first-pass result. We take the last pass's output as $\hat{\mathbf r}$ and use its acoustic scoring function $S$ in the margin check below.

\subsection{Margin check}
\label{ssec:margin}
The search may still leave the default on weak evidence. We therefore initialize $\tilde{\mathbf r}=\hat{\mathbf r}$ and revisit the changed spans from left to right.
With a hyperparameter $\tau\ge 0$, at each span $j$ we keep the current reading only if its score advantage over the default meets a length-scaled threshold:
\begin{equation}
 S\big(y(\tilde{\mathbf r}[j\!\leftarrow\! r_j^0]),x\big) -S\big(y(\tilde{\mathbf r}),x\big)\ge\tau\,\ell_j,
 \label{eq:gate}
\end{equation}
where $\ell_j=\max(1,|\tilde r_j|,|r_j^0|)$, with $|r|$ counting symbols in the normalized pronunciation alphabet; thus, $\tau$ is the required score gain per symbol.
If this condition is not met, the span is reverted, $\tilde{\mathbf r}\leftarrow \tilde{\mathbf r}[j\!\leftarrow\! r_j^0]$, before the next span is checked. Setting $\tau=0$ disables this check.

\section{Experiments}
\label{sec:setup}

\subsection{Japanese resources and protocol}
\label{ssec:japanese}

Japanese is a challenging language for acoustic disambiguation: one orthographic string can have several valid readings (\ja{明日}: \emph{asu}, \emph{ashita}, or \emph{my\=onichi}), and the goal is to select the reading actually spoken in the recording.

\noindent\textbf{Reading candidates.}
For reading candidate generation, we use Japanese morphological analyzers and G2P tools such as Sudachi~\cite{takaoka2018sudachi}, Open~JTalk\footnote{\url{https://open-jtalk.sp.nitech.ac.jp/}}, MeCab/UniDic~\cite{kudo2004mecab}, and rule-based variants.
Readings found only by isolated-span analysis are less reliable, so the penalty $P$ in Eq.~\eqref{eq:objective} is set to $0.03\,m\,n_{\mathrm{ctc}}$, where $m$ counts spans switched from the default to such a reading and $n_{\mathrm{ctc}}$ is the utterance length in token count of the main scorer $L_1$.

\noindent\textbf{Scorers.}
We use a 320M-parameter wav2vec2.0 kana CTC model~\cite{sakasegawa2025hiragana} as the first scorer and an 810M-parameter autoregressive kana-whisper~\cite{sarashina2026tts} as the second. Kana-whisper allows multiple spellings for a pronunciation (e.g., long vowels, \ja{は}/\ja{わ}), so $L_2$ is the negative logarithm of the teacher-forced likelihoods summed over up to 16 spellings, caching cross-attention per recording.

\noindent\textbf{Hyperparameters.}
We use second scorer weight $\lambda=1$ (from $\{0.5,1,2\}$) and margin check threshold $\tau=0.6$ (see Eq.~\ref{eq:gate}), with the margin check using the final $S$ and $\ell_j$ in normalized kana.
For calibration, we use 1,000 utterances each from JSUT and JVS-par, transferred to JVS-dev; the spelling limit is set on JVS-dev.
Reading candidate and prompt settings also use these calibration subsets, which are excluded from evaluation.

\noindent\textbf{Data and transcripts.}
We use JVS-dev (600 nonparallel utterances, 20 speakers, manual kana)~\cite{takamichi2019jvs}, JSUT BASIC5000 (5,000 utterances, one speaker, verified labels)~\cite{sonobe2017jsut}, and JVS-par (9,997 utterances, 100 speakers, alignment-derived silver labels)\footnote{\url{https://github.com/r9y9/jvs_r9y9}}, evaluating all systems on 512 common utterances per corpus. Transcripts are corpus text (reference) or Whisper large-v3 output~\cite{radford2023whisper} (ASR).

\noindent\textbf{Baselines and references} (Tab.~\ref{tab:main}).
We compare five categories of baselines:
\emph{(A) Text-only G2P:} Open~JTalk, Sudachi + Open~JTalk, and Sudachi + Open~JTalk with spoken-form rules (our default, used as the control for acoustic selection).
\emph{(B) Audio-only decoding:} greedy Kana CTC and kana-whisper with the same checkpoints as our scorers.
\emph{(C) Same-model text conditioning:} kana-whisper with a transcript prompt, and Kana CTC with dictionary-constrained greedy decoding inspired by~\cite{hien2025grapheme}, with and without a long-vowel lattice.
\emph{(D) Released ST2P annotator:} Furigana Whisper\footnote{\url{https://huggingface.co/Parakeet-Inc/furigana_whisper_small_jsut}}, a public audio-and-text kana model, with its published decoding and with the same dictionary constraint over our candidates (not a reproduction of the unavailable model of~\cite{hien2025grapheme}); JSUT results are omitted because it was trained on JSUT.
\emph{(E) Multimodal LLMs (MLLMs):} Qwen3-Omni-30B-A3B, Qwen2.5-Omni-7B, Gemma-3n-E4B, and Phi-4-multimodal~\cite{qwen3omni,qwen25omni,phi4mm}.
We report Gemini 2.5/3/3.5/3.6 Flash~\cite{geminiflash} as commercial references, which are not highlighted for best scores due to their high labeling cost. All MLLMs are given the same audio, transcripts, three fixed examples, prompt tuned by validation, and one output repair.

\noindent\textbf{Metric.}
We report corpus-level character error rate (CER, \%) after normalization (katakana, punctuation, long vowels, fixed mappings) to ignore notational differences. CER is computed on normalized kana, or on silver labels for JVS-par.

\begin{table}[t]
    \centering
    \footnotesize
    \setlength{\tabcolsep}{1pt}
    \caption{Japanese CER (\%, $\downarrow$) on 512 common utterances per corpus. dev/par: JVS-dev/JVS-par. $\dagger$: dictionary-constrained decoding inspired by~\cite{hien2025grapheme}. Furigana Whisper's JSUT cells are omitted (training data). Gray: commercial references. Bold: column minima excluding commercial models.}
    \label{tab:main}
    \vspace{2pt}
    \begin{tabular}{lrrrrrr}
    \toprule
    & \multicolumn{3}{c}{Reference text} & \multicolumn{3}{c}{ASR text}\\
    \cmidrule(lr){2-4}\cmidrule(lr){5-7}
    System & dev & JSUT & par & dev & JSUT & par\\
    \midrule
    \rowcolor{gray!15}
    \multicolumn{7}{l}{\emph{Commercial models (reference only): audio + text}}\\
    \rowcolor{gray!15}
    \quad Gemini 2.5 Flash & 0.62 & 0.84 & 1.29 & 0.98 & 1.14 & 1.95\\
    \rowcolor{gray!15}
    \quad Gemini 3 Flash & 0.64 & 0.52 & 0.74 & 0.88 & 0.68 & 1.72\\
    \rowcolor{gray!15}
    \quad Gemini 3.5 Flash & 0.50 & 0.59 & 0.84 & 0.82 & 0.69 & 1.67\\
    \rowcolor{gray!15}
    \quad Gemini 3.6 Flash & 1.19 & 1.69 & 1.17 & 1.95 & 2.55 & 1.87\\
    \midrule
    \multicolumn{7}{l}{\emph{(A) Text only: G2P}}\\
    \quad Open JTalk & 0.98 & 1.52 & 1.35 & 1.64 & 2.12 & 2.82\\
    \quad Sudachi + Open JTalk & 1.27 & 1.78 & 0.67 & 2.01 & 2.36 & 2.57\\
    \quad \quad + spoken-form rules & 0.85 & 1.40 & 0.60 & 1.53 & 1.97 & 2.49\\
    \midrule
    \multicolumn{7}{l}{\emph{(B) Audio only: same-model decoding}}\\
    \quad Kana CTC (greedy) & 5.38 & 5.61 & 13.02 & -- & -- & --\\
    \quad kana-whisper & 0.90 & 0.87 & 2.29 & -- & -- & --\\
    \midrule
    \multicolumn{7}{l}{\emph{(C) Audio + text: same-model conditioning}}\\
    \quad kana-whisper + prompt & 0.71 & 0.79 & 1.64 & 0.79 & 0.81 & 2.06\\
    \quad CTC + dict.\ constraint$^\dagger$ & 0.80 & 1.40 & 0.68 & 1.37 & 1.92 & 2.56\\
    \quad \quad + long-vowel lattice$^\dagger$ & 0.80 & 1.27 & 0.65 & 1.39 & 1.83 & 2.50\\
    \midrule
    \multicolumn{7}{l}{\emph{(D) Audio + text: released ST2P annotator}}\\
    \quad Furigana Whisper + prompt & 0.90 & -- & 0.63 & 1.21 & -- & 2.05\\
    \quad \quad + dict.\ constraint$^\dagger$ & 0.21 & -- & 0.15 & 0.73 & -- & 1.66\\
    \midrule
    \multicolumn{7}{l}{\emph{(E) Audio + text: open MLLMs}}\\
    \quad Qwen3-Omni & 8.13 & 10.31 & 10.03 & 8.47 & 11.31 & 8.09\\
    \quad Qwen2.5-Omni & 16.85 & 23.89 & 25.86 & 17.18 & 25.94 & 27.43\\
    \quad Gemma-3n-E4B & 14.97 & 18.30 & 13.99 & 13.89 & 17.69 & 15.61\\
    \quad Phi-4-multimodal & 61.48 & 65.44 & 68.92 & 58.10 & 64.70 & 74.02\\
    \midrule
    \multicolumn{7}{l}{\emph{Audio + text: proposed rescoring}}\\
    \quad Ours (AR-only) & 0.25 & 0.27 & 0.31 & 0.72 & 0.75 & 1.81\\
    \quad Ours (CTC-only) & 0.23 & 0.30 & 0.13 & 0.70 & 0.74 & 1.65\\
    \quad Ours (cascade) & \textbf{0.15} & \textbf{0.17} & \textbf{0.04} & \textbf{0.64} & \textbf{0.65} & \textbf{1.58}\\
    \bottomrule
    \end{tabular}
    \end{table}
\subsection{Japanese results}
\label{sec:results}

\noindent\textbf{Main results (Tab.~\ref{tab:main}).}
Our cascade outperforms all baselines in both the reference-text and ASR-text settings, including ST2P Furigana Whisper with dictionary constraint, which requires pronunciation-labeled speech; ours uses only frozen components.
It also substantially improves over greedy Kana CTC decoding (5.38/5.61/13.02 to 0.154/0.171/0.042) and the text-only default (0.85/1.40/0.60), showing that combining S2P scores and G2P candidates is more effective than either alone. The cascade also outperforms CTC-only and AR-only rescoring on all corpora.

\noindent\textbf{Ablations (Tab.~\ref{tab:abl}).}
We validate each component by removing the margin check (w/o gate, $\tau=0$), removing the cascade (w/o cascade, fusing every utterance), using beam search with width $B=5$ (w/ beam), or replacing the input audio with unrelated or silent audio (other audio, silent audio).
For reference, we also report a greedy oracle that selects candidates with minimum edit distance to the reference reading, an approximate oracle within the candidate set.

Tab.~\ref{tab:abl} shows that removing the gate or cascade raises CER by up to 0.022 and 0.016, indicating that both components are effective.
Replacing audio with other recordings or silence also increases CER to 12.4--16.2\%, confirming that acoustic evidence is necessary.
Finally, beam search performs similarly to greedy search, supporting that each span's acoustic evidence is mostly local. We thus prefer greedy search for its lower cost (Sec.~\ref{sec:cost}).

The cascade is only 0.107/0.050/0.042 points above the oracle, recovering 87--96\% of the oracle's CER reduction over text-only.

\begin{table}[t]
\centering
\footnotesize
\setlength{\tabcolsep}{4pt}
\caption{Japanese ablations (CER \%, reference text, 512 utterances). Proposed: cascade, gate ($\tau=0.6$), greedy search. w/o gate: $\tau=0$; w/o cascade: fusion every utterance; w/ beam: $B=5$. Other/silent audio: use unrelated or silent audio, keeping text/candidates.}
\label{tab:abl}
\vspace{2pt}
\begin{tabular}{lrrr}
\toprule
Configuration & JVS-dev & JSUT & JVS-par\\
\midrule
\multicolumn{4}{l}{\emph{Components}}\\
\quad w/o gate & \textbf{0.154} & 0.193 & 0.054\\
\quad w/o cascade & 0.161 & 0.182 & 0.058\\
\quad w/ beam & 0.167 & \textbf{0.160} & \textbf{0.042}\\
\midrule
\multicolumn{4}{l}{\emph{Audio substitution}}\\
\quad Proposed w/ other audio & 15.329 & 16.192 & 12.440\\
\quad Proposed w/ silent audio & 13.870 & 14.819 & 12.544\\
\midrule
\rowcolor[gray]{0.92}
\multicolumn{4}{l}{\emph{Oracle Reference}}\\
\rowcolor{gray!15}
\quad Greedy oracle & \emph{0.047} & \emph{0.121} & \emph{0.000}\\
\midrule
Proposed & \textbf{0.154} & 0.171 & \textbf{0.042}\\
\bottomrule
\end{tabular}
\end{table}

\subsection{Cost and accuracy (Fig.~\ref{fig:rtf})}
\label{sec:cost}

We measured real-time factor (RTF) for all audio-based systems on 512 JVS-dev utterances with reference transcripts, using a single H100 GPU (batch size one).

Fig.~\ref{fig:rtf} plots RTF against CER: the proposed methods achieve the lowest CER while remaining comparatively inexpensive. Greedy search is 3--3.5$\times$ faster than beam search at comparable CER (Tab.~\ref{tab:abl}).
The cascade method is also twice as fast as kana-whisper decoding because it runs kana-whisper only on selected spans and scores G2P candidates with a single forward pass instead of stepwise decoding.

\subsection{Multilingual evaluation}
\label{sec:multilingual}

We test transfer beyond Japanese with reference transcripts, IPA candidates, and language-specific resources.

\noindent\textbf{Candidates and scorers.}
\label{ssec:multi-method}
Spanish and French candidates come from CharsiuG2P~\cite{zhu2022charsiu} with eSpeak NG and Epitran as fallbacks; English uses the MFA \texttt{english\_us\_arpa} dictionary~\cite{mfadictionary}. Each pool is extended to up to 16 candidates with additional G2P outputs and local substitution or deletion of phonetically similar segments to cover unlisted variants. PhoneticXeus~\cite{bharadwaj2026phoneticxeus} is the main scorer, and fusion adds the POWSM CTC head~\cite{li2026powsm}, both scored via Eq.~\ref{eq:nll}.

We reuse the Japanese settings (Sec.~\ref{ssec:japanese}): greedy search, cascade with $\lambda=1$, and the margin check (Sec.~\ref{ssec:margin}) with $\ell_j$ counted in normalized IPA segments. The two main changes are: $P=0$, since these G2P and dictionary candidates contain no isolated-span readings that the Japanese penalty targets; and $\tau$ is set per language (2.0 for Spanish, 1.5 for French, 0 for English) according to each development set to minimize PFER.

\noindent\textbf{Data.}
We evaluate Spanish DIMEx100~\cite{pineda2004dimex}, French Rhapsodie~\cite{lacheret2014rhapsodie}, and English Buckeye~\cite{pitt2007buckeye}, sampling 200 utterances per language that contain at least one variable candidate span (192 for French after removing utterances without audio). Spanish and French use six speaker groups each for evaluation and configuration; English uses four groups each, all drawn from the corpus's designated development speakers, so its results are preliminary.

\noindent\textbf{Baselines and references} (Tab.~\ref{tab:multi}).
\emph{(A) Text-only specialist:} the default reading from the language's dictionary or G2P. \emph{(B) Open MLLMs:} models from Sec.~\ref{ssec:japanese}, evaluated alongside commercial references with the same audio, text, and prompting protocol.

\noindent\textbf{Metric.}
We report phonological feature error rate, $\mathrm{PFER}=100\sum\nolimits_i w_i\,d_F(y_i,\hat y_i)/\sum\nolimits_i w_i\,|y_i|$, where $w_i$ is an inverse-sampling weight (eligible-to-sampled ratio), and $d_F$ is an edit distance using PanPhon feature distances~\cite{mortensen2016panphon} for substitutions and unit costs for insertions and deletions. Fixed IPA mappings align conventions within each language. PFER scores are not directly comparable across languages.

\begin{table}[t]
\centering
\footnotesize
\setlength{\tabcolsep}{5pt}
\caption{PFER (\%, $\downarrow$) on shared subsets with candidate variation. (A): reference text only; others: text and audio. Gray: commercial references. Bold: best (non-commercial).}
\label{tab:multi}
\vspace{2pt}
\begin{tabular}{lrrr}
\toprule
System & Spanish & French & English\\
    & $n=200$ & $n=192$ & $n=200$\\
\midrule
\rowcolor{gray!15}
\multicolumn{4}{l}{\emph{Commercial models (reference only): audio + text}}\\
\rowcolor{gray!15}
\quad Gemini 2.5 Flash & 2.23 & 4.41 & 13.84\\
\rowcolor{gray!15}
\quad Gemini 3 Flash & 2.13 & 2.98 & 13.05\\
\rowcolor{gray!15}
\quad Gemini 3.5 Flash & 2.01 & 4.05 & 13.85\\
\rowcolor{gray!15}
\quad Gemini 3.6 Flash & 2.53 & 3.82 & 13.70\\
\midrule
\multicolumn{4}{l}{\emph{(A) Text only: dictionary / G2P}}\\
\quad Specialist default reading & 2.47 & 4.39 & 13.76\\
\midrule
\multicolumn{4}{l}{\emph{(B) Audio + text: open MLLMs}}\\
\quad Qwen3-Omni & 3.90 & 11.08 & 17.05\\
\quad Qwen2.5-Omni & 5.99 & 28.56 & 28.60\\
\quad Gemma-3n-E4B & 7.52 & 20.21 & 25.40\\
\quad Phi-4-multimodal & 57.32 & 128.35 & 74.38\\
\midrule
\multicolumn{4}{l}{\emph{Audio + text: proposed rescoring}}\\
\quad Ours (PhoneticXeus-only) & \textbf{2.38} & 4.33 & 14.01\\
\quad Ours (cascade) & 2.52 & \textbf{4.11} & \textbf{13.21}\\
\bottomrule
\end{tabular}
\end{table}

\noindent\textbf{Results (Tab.~\ref{tab:multi}).}
With only a change of lexical resources and scorers, and no training, the pipeline improves on the language-specific specialist in all three languages (2.47$\to$2.38 in Spanish with PhoneticXeus-only; 4.39$\to$4.11 and 13.76$\to$13.21 in French and English with the cascade). Both configurations outperform every open MLLM.
The best configuration is within 0.4 points of the top Gemini model in Spanish and English. Gains depend on resource and scorer matching, so choosing resources is key.

\balance
\section{Conclusion}
\label{sec:conclusion}
We proposed a training-free ST2P pipeline based on open models and lexical resources. For Japanese, it is faster and outperforms all baselines. The method also works for Spanish, French, and English.

\clearpage
\nobalance

\bibliographystyle{IEEEbib}
\bibliography{reference}

\end{document}